%% file: paper.tex
\documentclass[10pt,twocolumn,letterpaper]{article}

\usepackage[]{spconf} % To force page numbers, e.g. for an arXiv version

\usepackage{graphicx}
\usepackage{subcaption}
\usepackage{array}
\usepackage{float}
\usepackage{amsmath}

\usepackage{tikz}
\usetikzlibrary{positioning, shapes.geometric, arrows.meta, calc}

\tikzset{
    box/.style={rectangle, draw, rounded corners, minimum height=1.2em, minimum width=2em, align=center, font=\small},
    arrow/.style={-Stealth, thick},
    circ/.style={circle, draw, minimum size=2em, align=center, font=\small},
    labelbox/.style={rectangle, draw=none, align=center, font=\scriptsize},
    softbox/.style={rectangle, draw, fill=yellow!20, rounded corners, font=\small},
    augbox/.style={rectangle, draw, fill=green!20, rounded corners, font=\small},
    augbox2/.style={rectangle, draw, fill=blue!20, rounded corners, font=\small},
    lossbox/.style={rectangle, draw, fill=orange!30, rounded corners, font=\small}
}


\definecolor{wacvblue}{rgb}{0.21,0.49,0.74}
\usepackage[pagebackref,breaklinks,colorlinks,allcolors=wacvblue]{hyperref}

\title{ALDI-ray: Adapting the ALDI Framework for Security X-ray Object Detection}

\name{Omid Reza Heidari, Yang Wang, Xinxin Zuo}
\address{Concordia University, Montreal, Quebec, Canada}

\begin{document}
\maketitle
\input{sec/0_abstract}    
\input{sec/1_intro}
\input{sec/2_related_work}
\input{sec/3_method}
\input{sec/4_results_and_discussion}
\input{sec/5_conclusion}
\newpage
{
    \small
    \bibliographystyle{IEEEbib}
    \bibliography{main}
}

\end{document}

%% file: sec/0_abstract.tex
Domain adaptation in object detection is critical for real-world applications where distribution shifts degrade model performance. Security X-ray imaging presents a unique challenge due to variations in scanning devices and environmental conditions, leading to significant domain discrepancies. To address this, we apply ALDI++, a domain adaptation framework that integrates self-distillation, feature alignment, and enhanced training strategies to mitigate domain shift effectively in this area. We conduct extensive experiments on the EDS dataset, demonstrating that ALDI++ surpasses the state-of-the-art (SOTA) domain adaptation methods across multiple adaptation scenarios. In particular, ALDI++ with a  Vision Transformer for Detection (ViTDet) backbone achieves the highest mean average precision (mAP), confirming the effectiveness of transformer-based architectures for cross-domain object detection. Additionally, our category-wise analysis highlights consistent improvements in detection accuracy, reinforcing the robustness of the model across diverse object classes. Our findings establish ALDI++ as an efficient solution for domain-adaptive object detection, setting a new benchmark for performance stability and cross-domain generalization in security X-ray imagery.

%% file: sec/1_intro.tex
\section{Introduction}
\label{sec:intro}

Object detection in X-ray imagery plays a critical role in public safety domains such as airport security, customs inspection, and border control. The goal is to identify prohibited items such as knives, scissors, and power banks within cluttered and occluded luggage scenes. With the rise of deep learning, particularly Convolutional Neural Networks (CNNs), there has been substantial progress in improving detection accuracy and robustness for general object detection tasks~\cite{ Yin2021CenterPoint, Zhang2021Diversifying}. Transferring this success to the X-ray domain holds strong practical value, especially given the volume of security checks in high-throughput environments.

Existing X-ray detection approaches largely focus on improving network architectures or introducing task-specific modules. Prior works~\cite{Piao_Rong_Zhang_Lu_2020, Piao2020A2dele} have designed increasingly sophisticated models to handle overlapping objects and visual noise. However, these methods often overlook the domain shift problem. This shift arises due to variations in imaging devices, scanning geometry, or reconstruction techniques, which alter the underlying data distribution while maintaining similar appearance.

This issue is particularly evident in what is known as endogenous domain shift~\cite{Miao2019SIXray, Wang2021Towards}. These domain discrepancies are intrinsic to the data acquisition process and are not visually apparent to human observers. Inter-subset distribution shifts in the EDS dataset~\cite{PSN} significantly impact model performance. While previous works~\cite{Li_2020, tang2020unsuperviseddomainadaptationstructurally, wang2021domainspecificsuppressionadaptiveobject} in Unsupervised Domain Adaptation (UDA) have addressed cross-domain issues in natural images, few efforts have systematically addressed this challenge in security X-ray imagery. In this paper, we tackle the problem using ALDI++~\cite{ALDI}, which integrates feature alignment and self-distillation to enhance cross-domain generalization.

Our contributions are twofold: we identify the underexplored challenge of endogenous domain shift in X-ray object detection and show that ALDI++ effectively addresses it by surpassing both source-only baselines and state-of-the-art domain adaptation methods on the multi-domain EDS dataset~\cite{PSN}; furthermore, we are the first to adapt ALDI++ to this setting, validating its effectiveness through extensive evaluation, including per-domain and per-category analysis, ablation studies, and qualitative visualizations.

%% file: sec/2_related_work.tex
\section{Related Work}
\label{sec:related_work}
\textbf{Object Detection in X-ray Images}: Object detection is a core task in computer vision that involves identifying and localizing multiple objects within an image. Modern detectors typically followed either a one-stage or a two-stage design, balancing trade-offs between speed and accuracy. Recent advances in deep learning, particularly the introduction of transformer-based architectures such as ViTDet~\cite{li2022exploringplainvisiontransformer}, have further improved detection performance, especially for complex scenes with small or overlapping objects.

% While significant progress has been made on benchmark datasets, real-world deployment often introduces challenges such as occlusion, clutter, and variation in object appearance. These issues make object detection a persistently difficult problem, particularly in specialized domains like security or medical imaging, where data distributions may differ from standard training sets.

\noindent\textbf{Domain Shift in Object Detection}: Real-world object detectors often suffer significant performance degradation when applied to new environments due to domain shift. In UDA, a model that is trained on a labeled source domain must generalize to a different but related target domain without access to target annotations~\cite{oza2021unsuperviseddomainadaptationobject}. Even if object appearances remain consistent, variations in background, lighting, sensor characteristics, or imaging processes can subtly distort bounding-box predictions and classification scores. Since annotating target domain data is prohibitively expensive for every new domain, UDA methods must effectively leverage unlabeled target data. Recent surveys on Domain Adaptive Object Detection (DAOD)~\cite{oza2021unsuperviseddomainadaptationobject, li2020deepdomainadaptiveobject} showed that most methods combine image- or instance-level feature alignment with self-training on target data to mitigate cross-domain shifts. In response to this challenge, a variety of techniques have been proposed. For instance, Chen et al.~\cite{chen2018domainadaptivefasterrcnn} made significant strides in DAOD by aligning both image- and instance-level features using a domain-adversarial approach. Subsequently, further advancements were made by several works~\cite{saito2019strongweakdistributionalignmentadaptive, zhao2020collaborativetrainingregionproposal, zheng2020crossdomainobjectdetectioncoarsetofine, xu2020crossdomaindetectiongraphinducedprototype}, which introduced new strategies to improve adaptation.

\noindent\textbf{Domain Shift in X-ray Object Detection}: A research~\cite{zhao2020collaborativetrainingregionproposal} used collaborative training between region proposal and classification, while another research~\cite{zheng2020crossdomainobjectdetectioncoarsetofine} employed a coarse-to-fine alignment strategy to progressively reduce domain gaps. Most recently, Perturbation Suppression Network (PSN)~\cite{PSN} with a tailored benchmark has been introduced to study endogenous domain shift in X-ray images.

%% file: sec/3_method.tex
\section{Method}
\label{sec:method}
\textbf{Problem Formulation}: Given a labeled source domain \( \mathcal{D}_s = \{(x^s_i, y^s_i)\}_{i=1}^{N_s} \) and an unlabeled target domain \( \mathcal{D}_t = \{x^t_j\}_{j=1}^{N_t} \), the goal is to train an object detector that performs well on \( \mathcal{D}_t \), despite the domain shift between \( \mathcal{D}_s \) and \( \mathcal{D}_t \). This setting is known as unsupervised domain adaptation for object detection, where the key challenge is to generalize detection performance from \( \mathcal{D}_s \) to \( \mathcal{D}_t \) without access to target labels.

\noindent\textbf{Baseline: Faster R-CNN}: Faster R-CNN~\cite{ren2016fasterrcnnrealtimeobject}
is a two-stage object detector that serves as a standard baseline in many vision tasks. It first generates candidate object regions via a Region Proposal Network (RPN) and then classifies and refines these proposals with a second-stage classifier/regressor that shares the backbone features. We adopt a vanilla Faster R-CNN with a VGG-16 backbone as our source-only detector.

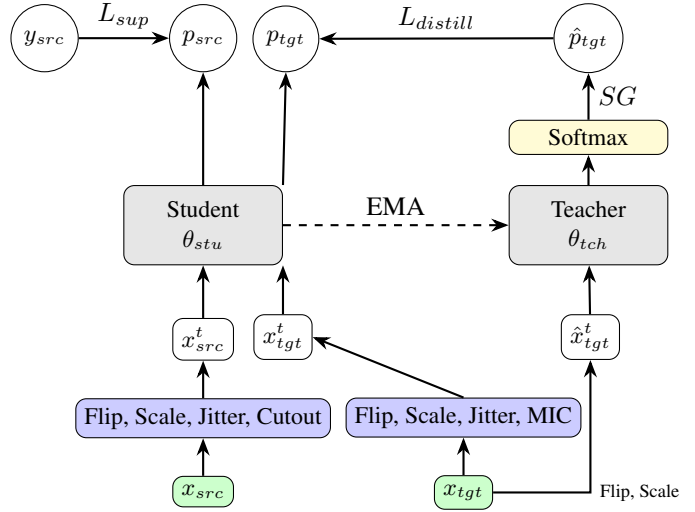
\begin{figure}
  \centering
    \begin{tikzpicture}[node distance=1.2cm and 1.2cm]

% Source side
\node[box, minimum width=6em, minimum height=3em, fill=gray!20] (student) {Student\\$\theta_{stu}$};
\node[circ, above=1.5cm of student] (psrc) {$p_{src}$};
\node[circ, left=of psrc] (ysrc) {$y_{src}$};
\node[box, below=0.7cm of student] (xsrc_t) {$x^{t}_{src}$};
\node[augbox2, below=0.5cm of xsrc_t] (xsource) {Flip, Scale, Jitter, Cutout};
\node[augbox, below=0.5cm of xsource] (xsrc) {$x_{src}$};

\draw[arrow] (xsrc) -- (xsource);
\draw[arrow] (xsource) -- (xsrc_t);
\draw[arrow] (xsrc_t) -- (student.south);
\draw[arrow] (ysrc) -- node[above] {$L_{sup}$} (psrc);
\draw[arrow] (student.north) -- (psrc);

% Target side
\node[box, below=0.65cm of student.south east] (xtgt_t) {$x^{t}_{tgt}$};
\node[augbox2, below=0.5cm of xtgt_t, xshift=2.4cm] (xtgt) {Flip, Scale, Jitter, MIC};
\node[augbox, below=0.5cm of xtgt] (xtarget) {$x_{tgt}$};
\draw[arrow] (xtarget) -- (xtgt);
\draw[arrow] (xtgt.north) -- (xtgt_t.east);
\draw[arrow] (xtgt_t) -- (student.south east);
\node[circ, above=1.5cm of student, xshift=1.1cm] (ptgt) {$p_{tgt}$};
\draw[arrow] (student.north east) -- (ptgt.south);

% Teacher side
\node[box, right=of xtgt_t, xshift=2.1cm] (xtgt_hat) {$\hat{x}^{t}_{tgt}$};
\node[box, right=3cm of student, fill=gray!20, minimum width=6em, minimum 
height=3em] (teacher) {Teacher\\$\theta_{tch}$};
\node[softbox, above=0.4cm of teacher, minimum width=6em] (softmax) {Softmax};
\node[circ, above=0.62cm of softmax] (ptgt_hat) {$\hat{p}_{tgt}$};
\draw[arrow] (teacher) -- (softmax);
\draw[arrow] (softmax) -- node[right] {$SG$} (ptgt_hat);
\draw[arrow] (ptgt_hat) -- node[above] {$L_{distill}$} (ptgt);
\draw[arrow] (xtarget.east) -| node[right, font=\scriptsize] {Flip, Scale} (xtgt_hat);
\draw[arrow] (xtgt_hat) -- (teacher);

% EMA Arrow
\draw[arrow, dashed] (student.east) -- node[above] {EMA} (teacher.west);

\end{tikzpicture}
    \caption{An overview of the ALDI++ training pipeline.}
  \label{fig:aldi_archit}
\end{figure}

\noindent\textbf{Model}: With respect to the selection of the model, we utilize ALDI++~\cite{ALDI}, due to several justifications and advantages. Firstly, it represents the best performance for domain adaptation for the Cityscapes $\rightarrow$ Foggy Cityscapes benchmark~\cite{cordts2016cityscapesdatasetsemanticurban}, which is widely used in the previous works. It, moreover, has a simplified design compared to other available architectures. For instance, UMT uses a separate image-to-image translation network to align domains~\cite{ALDI}. The architecture, as shown in Figure~\ref{fig:aldi_archit}, consists of three main components, described as follows. 

\textit{Supervised Training with Source Domain Data}: For each annotated source sample \( x_{\text{src},i} \), a transformation \( t \sim T_{\text{src}} \) is applied, where \( T_{\text{src}} \) denotes the collection of permissible transformations in the source domain. The resulting transformed sample is subsequently fed into the student model to compute the supervised loss \( L_{\text{sup}} \), based on the corresponding ground truth labels \( y_{\text{src},i} \). The supervised loss is formulated using \( L(\cdot, \cdot) \), which represents standard object detection loss functions, such as those used in Faster R-CNN~\cite{ren2016fasterrcnnrealtimeobject}.

\textit{Self-Distillation with Target Domain Data}: Each unlabeled target sample \( x_{tgt,i} \) undergoes two distinct augmentation strategies. A weak transformation, \( \hat{t} \sim T_{\text{weak}} \), is applied to the teacher model, while a stronger transformation, \( t \sim T_{\text{tgt}} \), is used for the student model~\cite{cardace2022selfdistillationunsupervised3ddomain, ALDI}. The teacher model generates predictions, which serve as distillation targets for the student model’s predictions, \( p_{tgt,i} \)~\cite{french2018selfensemblingvisualdomainadaptation}. This framework provides a unified formulation for various knowledge distillation techniques, enabling flexibility in adapting different approaches within a consistent optimization objective~\cite{xie2020selftrainingnoisystudentimproves}.

\textit{Feature Alignment}: To enhance domain invariance, the source samples $x_{\text{src},i}$ and target samples $x_{\text{tgt},i}$ can be optionally aligned through an alignment objective $\mathcal{L}_{\text{align}}$. This objective facilitates consistency across domains at either the image or feature level. While this formulation remains general, our study specifically considers two widely used alignment techniques: domain-adversarial training and image-to-image alignment.

% Domain-adversarial training, exemplified by the Domain-Adversarial Neural Network (DANN)~\cite{ganin2015unsuperviseddomainadaptationbackpropagation}, employs a domain classifier $D$ to differentiate between source and target features. Simultaneously, the feature extractor is trained adversarially to prevent $D$ from making accurate domain distinctions.

% Alternatively, image-to-image alignment seeks to achieve domain invariance at the pixel level~\cite{bousmalis2017unsupervisedpixelleveldomainadaptation}. Utilizing generative models such as CycleGAN~\cite{8237506}, images are transformed between source and target domains to mitigate domain discrepancies~\cite{hoffman2017cycadacycleconsistentadversarialdomain}.

\textit{From ALDI to ALDI++}: ALDI++ is an improved version of the ALDI framework, incorporating several enhancements to optimize domain adaptation in object detection. These modifications improve the stability and effectiveness of the approach.

To begin with, a robust burn-in strategy is employed to enhance pseudo-label quality in the early stages of self-training. Since initial pseudo-label reliability depends on the out-of-distribution (OOD) generalization of the teacher model, a pre-training phase is implemented using strong data augmentations along with an Exponential Moving Average (EMA) copy of the model. This pre-training phase improves OOD generalization and accelerates convergence.

In addition to that, multi-task soft distillation replaces hard pseudo-labeling with soft distillation losses inspired by knowledge distillation literature. Instead of applying confidence thresholding to teacher predictions, unmodified teacher outputs serve as distillation targets, eliminating sensitivity to confidence thresholds. This technique enables direct distillation of multiple Faster R-CNN components, including RPN localization and objectness, as well as Region-of-Interest (RoI) classification and localization, leading to more stable and effective self-training.

Finally, training strategies for DAOD are re-evaluated to establish stronger baselines for ALDI++. Two key modifications consistently improve adaptation: (1) applying strong regularization to both source and target data and (2) ensuring balanced supervision by maintaining equal batch sizes from both domains. Additionally, feature alignment is disabled to enhance training stability without significantly impacting accuracy. These refinements collectively strengthen ALDI++, making it a more robust and generalizable approach for domain adaptation in object detection.

%% file: sec/4_results_and_discussion.tex
\section{Experimental Results}
\label{sec:result_discussion}

% \subsection{Dataset}
% In terms of available security X-ray images, there are various datasets such as EDS\cite{PSN} and CLCXray~\cite{CLCXray}. However, only a few of them can be utilized for the domain adaptation task due to certain restrictions such as class consistency and environmental similarity. Therefore, we select the EDS dataset for several justifications. First and foremost, it contains three datasets, each collected during different periods, which makes the comparison more reliable. In addition, each dataset has approximately the same number of samples. Furthermore, the categories are balanced, reducing the risk of overfitting.

% Pertaining to the structure of the dataset, it comprises 3 sub-datasets, namely EDS1, EDS2, and EDS3, each containing the same ten distinct classifications.  The sub-datasets are evenly distributed across ten distinct classes, ensuring a balanced representation. Each class appears approximately between 1,000 and 1,100 times in each dataset, except for four classes: umbrella, device, glass bottle, and power bank, in EDS1.

\textbf{Implementation Details}: % To ensure a fair comparison and stable training process, we maintain consistent hyperparameter settings across all experiments. The key parameters are summarized in Table~\ref{tab:hyperparams}.

% An exception is the learning rate for the VGG-16 backbone, where a higher value (0.06) was selected based on common practice in prior research. This adjustment aligns with established configurations for training shallow convolutional architectures and ensures the VGG-based models converge effectively under comparable training conditions.

\textit{Backbone}: With regard to backbone selection, three backbones were selected. Initially, VGG-16 ~\cite{simonyan2015deepconvolutionalnetworkslargescale} is used to ensure a fair comparison with existing methods, as well as the source-only model. In addition to VGG-16, Feature Pyramid Network (FPN)~\cite{lin2017featurepyramidnetworksobject} is employed due to its ability to enhance multi-scale feature representation, which is crucial for detecting objects of varying sizes. Additionally, we adopt ViTDet~\cite{li2022exploringplainvisiontransformer} as the backbone due to its superior feature representation capabilities, which demonstrates state-of-the-art performance in object detection tasks. The transformer-based architecture effectively captures long-range dependencies, making it well-suited for DAOD scenarios.

\textit{Metrics}: We utilized mAP~\cite{ MAPEveringham2014ThePV}, the most common evaluation metric in object detection task, to assess overall performance. For per-category evaluation, we use Average Precision (AP)~\cite{AP}, which measures detection quality across different IoU thresholds.

\begin{figure}
  \centering
  \hfill
  \includegraphics[width=\linewidth]{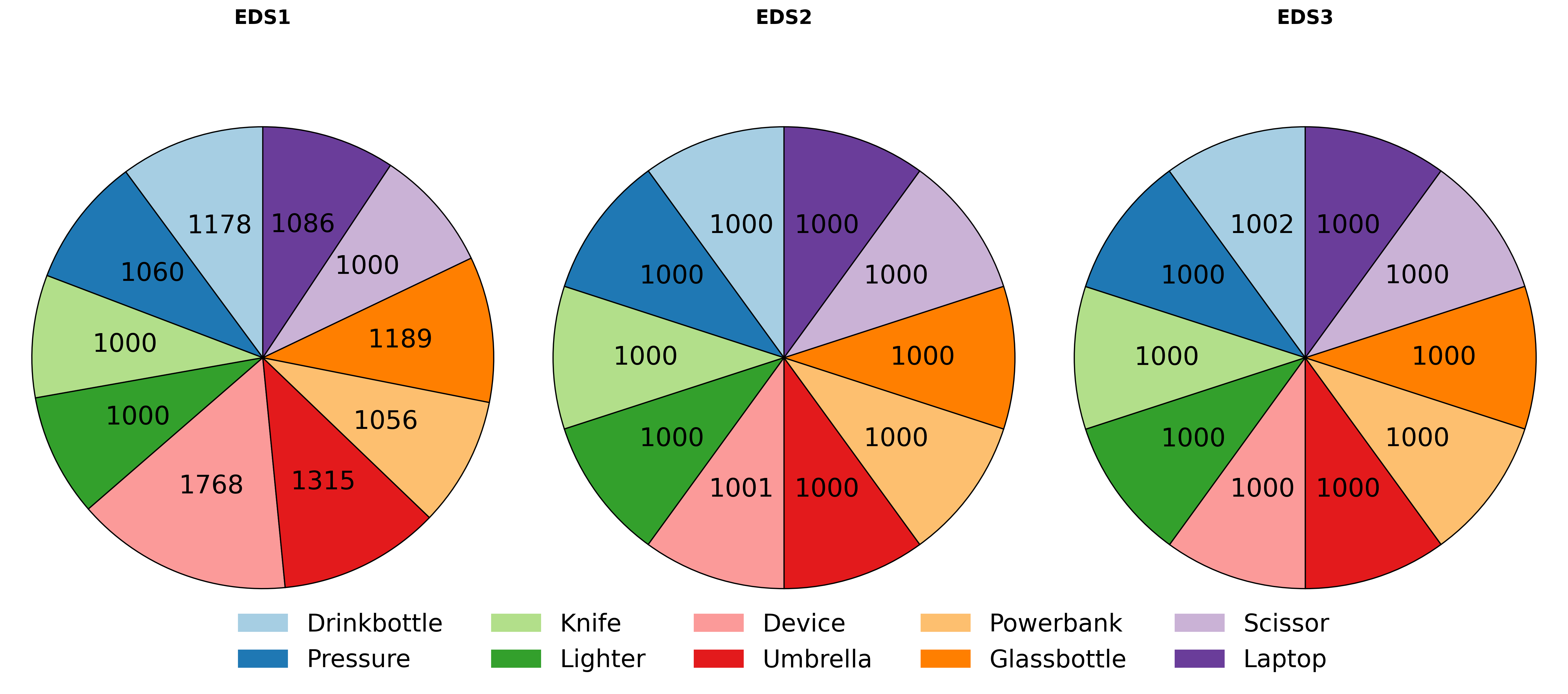}
  \caption{Distribution of the number of instances per object category in the EDS1, EDS2, and EDS3 sub-datasets. Each bar represents the frequency of a specific category within each domain, highlighting dataset balance and inter-domain variations.}
  \label{fig:eds_statistics}
\end{figure}

\textit{Dataset}: In terms of available security X-ray images, there are various datasets such as EDS~\cite{PSN} and CLCXray~\cite{CLCXray}. However, only a few of them can be utilized for the domain adaptation task due to certain restrictions such as class consistency and environmental similarity. Therefore, we select the EDS dataset for several justifications. Primarily, it contains three datasets, each collected during different periods, which makes the comparison more reliable. In addition, each dataset has approximately the same number of samples. Furthermore, the categories are balanced, reducing the risk of overfitting. More information regarding the EDS dataset can be observed in Figure~\ref{fig:eds_statistics}.

% \begin{table}[h]
% \centering
% \resizebox{\columnwidth}{!}{%
% \begin{tabular}{|c|c|}
% \hline
% \textbf{Hyperparameter} & \textbf{Value} \\
% \hline
% VGG-16 learning rate & 0.06 (Base), 0.04 (ALDI) \\ \hline
% FPN \& ViTDet learning rate & 0.0001 (Base), 0.04 (ALDI) \\ \hline
% Optimizer & ADAMW~\cite{loshchilov2019decoupledweightdecayregularization} \\ \hline
% Batch size & 16 (source), 16 (target) \\ \hline
% EMA decay & 0.999 \\ \hline
% IoU threshold & 0.5 \\ \hline
% Transformations & Flip, Scale, SimCLR augs \\ \hline
% Backbones & VGG-16, FPN, ViTDet \\ \hline
% Epochs per phase & 30 \\ \hline
% \end{tabular}
% }
% \caption{Summary of training hyperparameters that are used in our experiments.}
% \label{tab:hyperparams}
% \end{table}

\begin{figure*}[t]
  \centering
  % --------- Success: D2->D3 ------------
  \begin{subfigure}[t]{0.18\linewidth}
        \fcolorbox{black}{white}{\includegraphics[width=\linewidth, height=2.5cm]{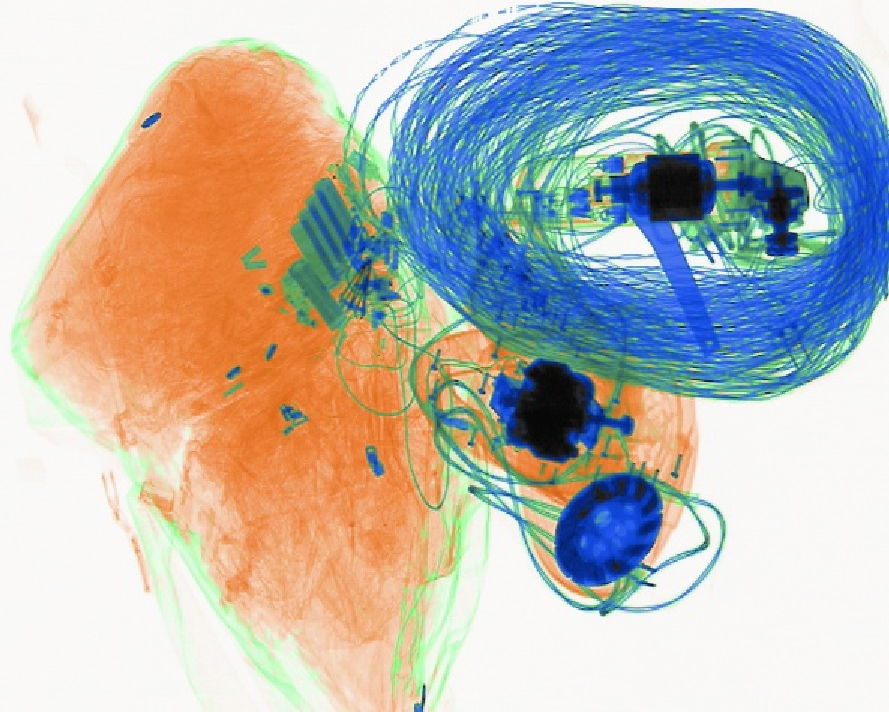}}
        \subcaption{}
  \end{subfigure}\hfill
  \begin{subfigure}[t]{0.18\linewidth}
        \fcolorbox{black}{white}{\includegraphics[width=\linewidth, height=2.5cm]{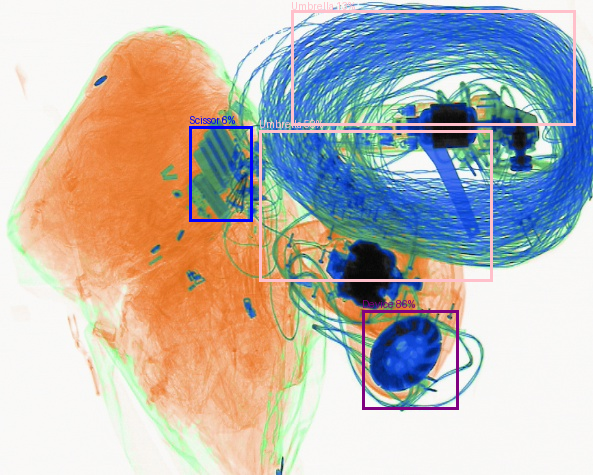}}
        \subcaption{}
  \end{subfigure}\hfill
  \begin{subfigure}[t]{0.18\linewidth}
        \fcolorbox{black}{white}{\includegraphics[width=\linewidth, height=2.5cm]{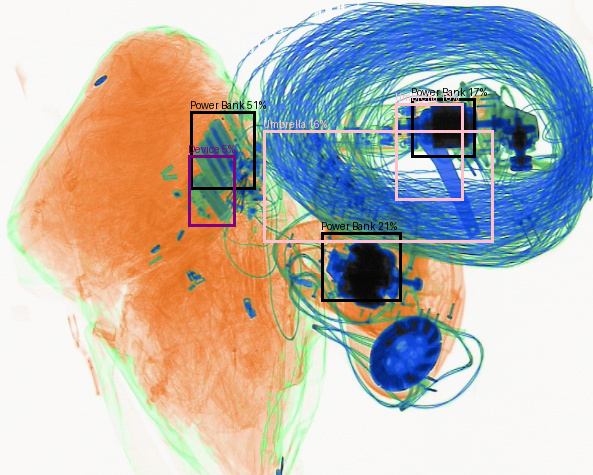}}
        \subcaption{}
  \end{subfigure}\hfill
  \begin{subfigure}[t]{0.18\linewidth}
        \fcolorbox{black}{white}{\includegraphics[width=\linewidth, height=2.5cm]{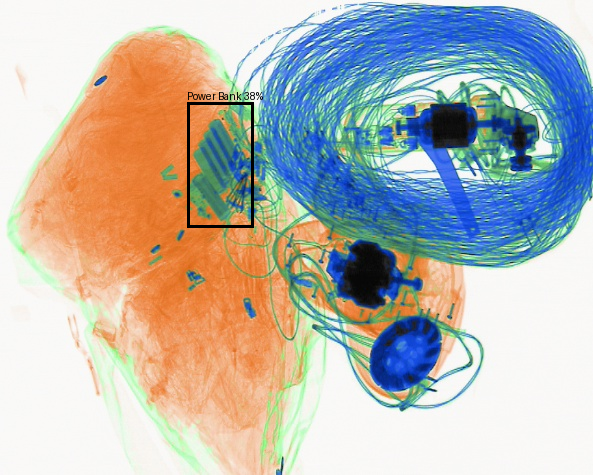}}
        \subcaption{}
  \end{subfigure}\hfill
  \begin{subfigure}[t]{0.18\linewidth}
        \fcolorbox{black}{white}{\includegraphics[width=\linewidth, height=2.5cm]{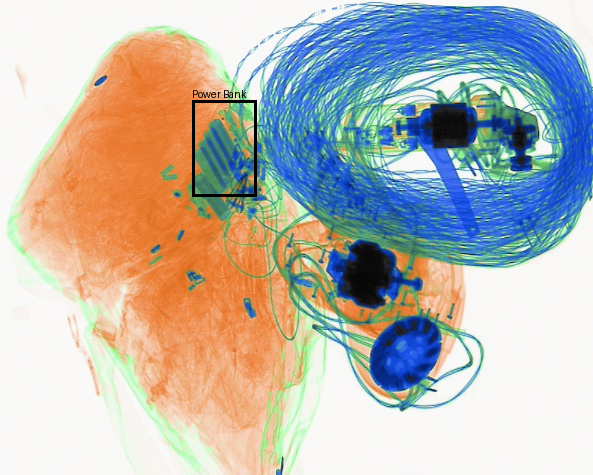}}
        \subcaption{}
  \end{subfigure}\hfill

  % --------- Success: D3->D1 ------------
  \begin{subfigure}[t]{0.18\linewidth}
        \fcolorbox{black}{white}{\includegraphics[width=\linewidth, height=2.5cm]{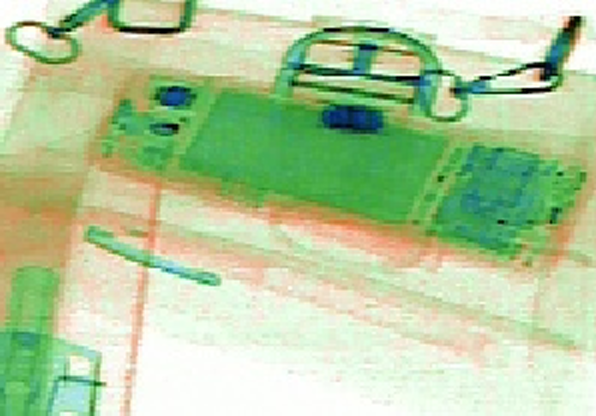}}
        \subcaption{}
  \end{subfigure}\hfill
  \begin{subfigure}[t]{0.18\linewidth}
        \fcolorbox{black}{white}{\includegraphics[width=\linewidth, height=2.5cm]{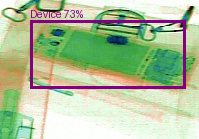}}
        \subcaption{}
  \end{subfigure}\hfill
  \begin{subfigure}[t]{0.18\linewidth}
        \fcolorbox{black}{white}{\includegraphics[width=\linewidth, height=2.5cm]{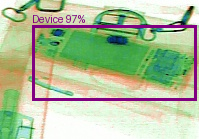}}
        \subcaption{}
  \end{subfigure}\hfill
  \begin{subfigure}[t]{0.18\linewidth}
        \fcolorbox{black}{white}{\includegraphics[width=\linewidth, height=2.5cm]{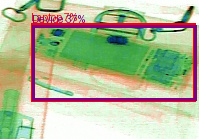}}
        \subcaption{}
  \end{subfigure}\hfill
  \begin{subfigure}[t]{0.18\linewidth}
        \fcolorbox{black}{white}{\includegraphics[width=\linewidth, height=2.5cm]{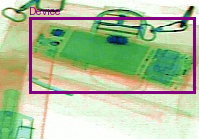}}
        \subcaption{}
  \end{subfigure}\hfill

  % --------- Failure: D2->D3 ------------
  \begin{subfigure}[t]{0.18\linewidth}
        \fcolorbox{black}{white}{\includegraphics[width=\linewidth, height=2.5cm]{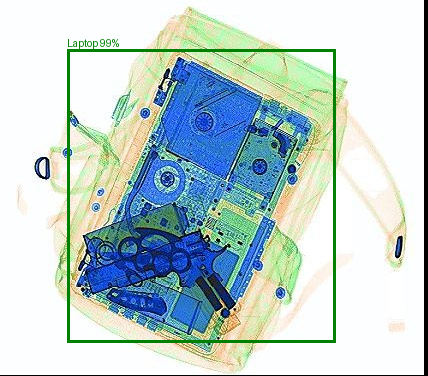}}
        \subcaption{}
  \end{subfigure}\hfill
  \begin{subfigure}[t]{0.18\linewidth}
        \fcolorbox{black}{white}{\includegraphics[width=\linewidth, height=2.5cm]{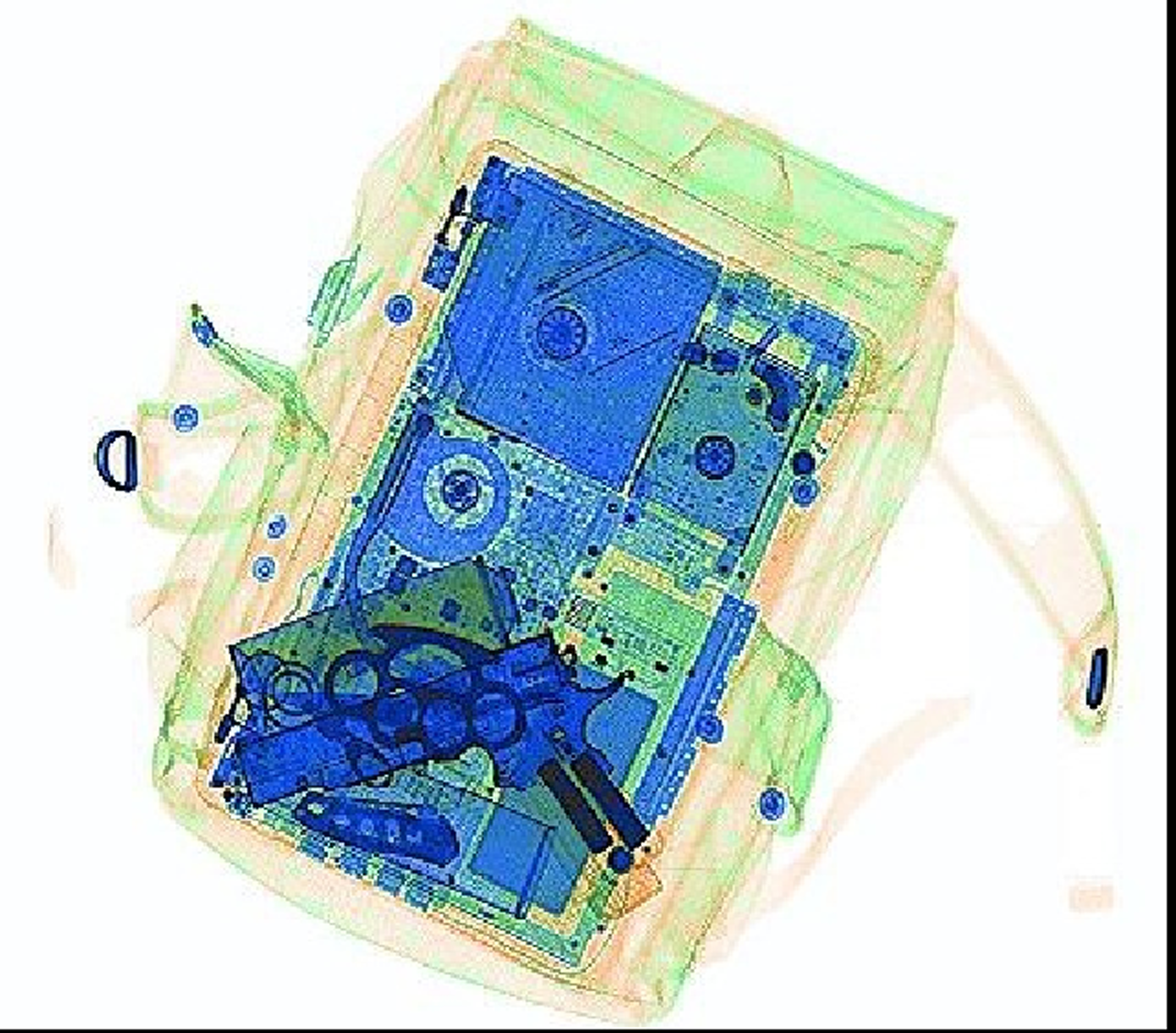}}
        \subcaption{}
  \end{subfigure}\hfill
  \begin{subfigure}[t]{0.18\linewidth}
        \fcolorbox{black}{white}{\includegraphics[width=\linewidth, height=2.5cm]{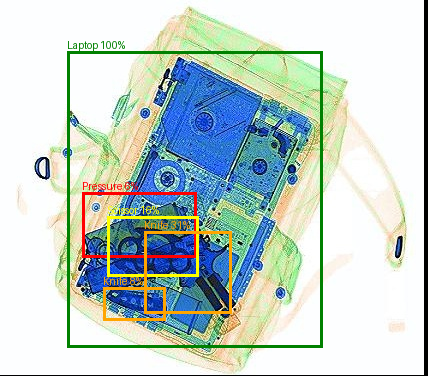}}
        \subcaption{}
  \end{subfigure}\hfill
  \begin{subfigure}[t]{0.18\linewidth}
        \fcolorbox{black}{white}{\includegraphics[width=\linewidth, height=2.5cm]{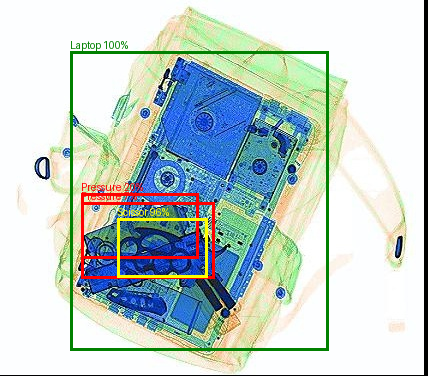}}
        \subcaption{}
  \end{subfigure}\hfill
  \begin{subfigure}[t]{0.18\linewidth}
        \fcolorbox{black}{white}{\includegraphics[width=\linewidth, height=2.5cm]{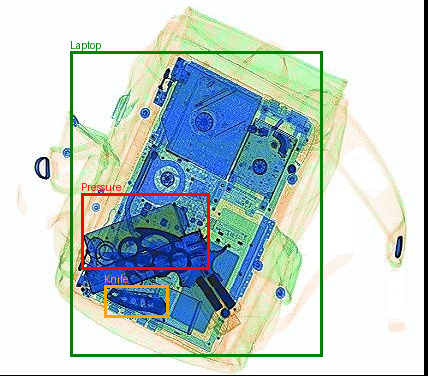}}
        \subcaption{}
  \end{subfigure}\hfill

  % --------- Caption ------------
  \caption{Qualitative comparison of detection results on domain adaptation tasks. 
  The \textbf{top two rows (a–j)} show successful detection cases: 
  (a–e) correspond to the D2$\rightarrow$D3 task and 
  (f–j) correspond to the D3$\rightarrow$D1 task. 
  The \textbf{bottom row (k–o)} illustrates a failure case from the D2$\rightarrow$D3 task.  
  For each group: (a, f, k) Source-only model; 
  (b, g, l) ALDI++ using VGG-16; 
  (c, h, m) ALDI++ using FPN; 
  (d, i, n) ALDI++ using ViTDet; 
  (e, j, o) Ground Truth.}
  \label{fig:qualitative_all}
\end{figure*}

\noindent\textbf{Qualitative Results}:
To illustrate the effectiveness of ALDI++ in mitigating domain shift, we present representative qualitative examples in Figure~\ref{fig:qualitative_all}. As shown in Figure~\ref{fig:qualitative_all}, source-only models often fail to detect or localize objects accurately, particularly under occlusion or clutter. In contrast, ALDI++ consistently produces more precise bounding boxes and higher confidence scores, with notable improvements for challenging categories such as \textit{knife} and \textit{glass bottle}.  

The comparison across backbone architectures further highlights the benefits of transformer-based models: while VGG-16 and FPN offer incremental gains, the ViTDet backbone achieves the most robust localization. Nonetheless, failure cases remain (Figure~\ref{fig:qualitative_all}), where all models struggle with visually ambiguous items such as \textit{knife}. These examples underscore both the strengths of ALDI++ in enhancing cross-domain generalization and the persistent challenges posed by subtle or occluded categories.

\noindent\textbf{Quantitative Results}

\textit{Per-Domain Analysis}: Regarding the mAP results on the EDS dataset, Table~\ref{table:overal_performance} shows that the ALDI++ architecture significantly outperforms the source-only model. Additionally, ALDI++ also outperforms existing methods across all settings, except when EDS dataset 2 is utilized as the target domain. Importantly, the figures of existing methods reported in Table~\ref{table:overal_performance} and Table~\ref{table:each_category} are sourced from the SOTA paper~\cite{PSN}. Furthermore, Table~\ref{table:overal_performance_differnt_backbone} presents a comparative analysis of different backbone architectures integrated with the ALDI++ framework.

\begin{table}[h]
\centering
\resizebox{\columnwidth}{!}{%
\begin{tabular}{|c | c | c | c | c | c | c | c|} 
\hline
 Method & $D_{1\rightarrow2}$ & $D_{1\rightarrow3}$ & $D_{2\rightarrow1}$ & $D_{2\rightarrow3}$ & $D_{3\rightarrow1}$ & $D_{3\rightarrow2}$ & Overall Avg.\\ [0.5ex] 
 \hline
  SO~\cite{ren2016fasterrcnnrealtimeobject} & 30.94	& 42.00 & 35.16 &	45.79 & 46.09 & 41.14 & 40.19\\ \hline
  DA~\cite{chen2018domainadaptivefasterrcnn}& 46.3	&55.6	&45.0	&57.5	&56.1&	54.6& 52.52\\ \hline
  SWDA~\cite{saito2019strongweakdistributionalignmentadaptive}& 46.9	&56.5	&49.7&	56.7	&56.6&	54.8& 53.53\\ \hline
  CST~\cite{zhao2020collaborativetrainingregionproposal}& 46.9	&54.3	&49.2	&55.5	&56.5	&52.8& 52.53\\ \hline
  CFA~\cite{zheng2020crossdomainobjectdetectioncoarsetofine} & 44.3&	53.7&	51.3	&53.6	&55.4	&51.3& 51.60\\ \hline
  PSN~\cite{PSN} & \textbf{48.3} &	57.6	&51.4	&57.8	&58.6	& \textbf{54.9}& 54.77\\ \hline
  ALDI++ (VGG-16)& 47.705 & \textbf{59.681} & \textbf{52.719} & \textbf{61.142} & \textbf{64.536} & 52.618& \textbf{56.4}\\ \hline
\end{tabular}
}
\caption{The mAP scores (\%) for various architectures across different source-target combinations in the EDS dataset.
For each $D_{S\rightarrow T}$, S and T refer to the source and target dataset, respectively. The results of ALDI++ with a VGG-16 backbone are compared with multiple existing domain adaptation methods. Note that the "Overall Avg." column summarizes average mAP across all transfer directions.
}
\label{table:overal_performance}
\end{table}

\begin{table}[h]
\centering
\resizebox{\columnwidth}{!}{%
\begin{tabular}{|c | c | c | c | c | c | c|} 
\hline
 Backbone & $D_{1\rightarrow2}$ & $D_{1\rightarrow3}$ & $D_{2\rightarrow1}$ & $D_{2\rightarrow3}$ & $D_{3\rightarrow1}$ & $D_{3\rightarrow2}$\\ [0.5ex] 
 \hline
  VGG-16& 47.705 & 59.681 & 52.719 & 61.142 & 64.536 & 52.618\\ \hline
  FPN& 59.594 & 73.493 & 64.432 & 71.284 & 73.759 & 66.131\\ \hline
  ViTDet& \textbf{61.522} & \textbf{76.115} & \textbf{68.919} & \textbf{79.432} & \textbf{77.257} & \textbf{71.92}\\ \hline
\end{tabular}
}
\caption{The mAP results (\%) for ALDI++ Architecture with different backbones across different source-target combinations in the EDS dataset.}
\label{table:overal_performance_differnt_backbone}
\end{table}
\textit{Per-Category Analysis}:
With respect to the AP scores for each object in the dataset, Table~\ref{table:each_category} reveals that ALDI++ yields the best results in all categories except LA and UM. Moreover, it reveals that, despite some categories such as PR and SC showing considerable improvements, the UM category did not exhibit any improvement following the application of ALDI++. The observed fluctuation in the enhancement of different categories can be attributed to the varying extent of the changes that occurred across the different datasets.

\begin{table}[h]
\centering
\resizebox{\columnwidth}{!}{%
\begin{tabular}{|c | c | c | c | c | c | c | c | c | c | c|} 
\hline
 Method & DB & PR & LI & KN & SE & PB & UM &  GB & SC & LA\\ [0.5ex] 
 \hline
  SO~\cite{ren2016fasterrcnnrealtimeobject} & 55.3 & 44.8 & 34.1 & 16.5 & 43.2 & 65.8 & 85.2 & 37.6 & 26.5 & 87.4 \\ \hline
  DA~\cite{chen2018domainadaptivefasterrcnn}& 54.7 & 52.7 & 38.6 & 15.4 & 47.7 & 68.3 & 86.7 & 40.2 & 30.2 & 90.1 \\ \hline
  SWDA~\cite{saito2019strongweakdistributionalignmentadaptive} & 55.6 & 52.6 & 40.9 & 17.3 & 49.5 & 69.8 & 86.7 & 41.1 & 30.0 & 90.3 \\ \hline
  CST~\cite{zhao2020collaborativetrainingregionproposal} & 55.1 & 51.2 & 39.0 & 16.0 & 49.6 & 69.5 & 86.5 & 40.7 & 25.0 & \textbf{92.2} \\ \hline
  CFA~\cite{zheng2020crossdomainobjectdetectioncoarsetofine} & 51.9 & 52.0 & 33.7 & 14.8 & 49.6 & 68.9 & 85.4 & 41.8 & 26.8 & 90.5 \\ \hline
  PSN~\cite{PSN} & 56.2 & 54.0 & 41.3 & 18.2 & 52.4 & 72.1 & \textbf{86.8} & 44.4 & 31.4 & 91.4 \\ \hline
  ALDI++ (VGG-16)& \textbf{58.25} & \textbf{67.41} & \textbf{43.29} & \textbf{41.37} & \textbf{58.77} & \textbf{73.1} & 77.13 & \textbf{53.19} & \textbf{56.82} & 90.46 \\ \hline
\end{tabular}
}
\caption{The AP scores (\%) for various architectures across different object categories in the EDS dataset.}
\label{table:each_category}
\end{table}

%% file: sec/5_conclusion.tex
\section{Conclusion}
\label{sec:conclusion}
This study investigates the application of ALDI++ for domain adaptation in object detection, with a focus on security X-ray imagery. Extensive experiments on the EDS dataset demonstrate that ALDI++ consistently outperforms both the source-only model and state-of-the-art domain adaptation methods, achieving superior cross-domain generalization. A category-wise analysis further confirms its effectiveness, with substantial improvements observed across most object classes, particularly PR, SC, and GB, which exhibit notable AP gains. Nevertheless, certain categories such as LA and DB show only limited improvement, indicating that specific object characteristics or domain discrepancies pose greater challenges for adaptation.